\documentclass[sigconf, 10pt]{acmart}
\usepackage{balance}
\usepackage{subcaption}
\usepackage{xcolor}
\usepackage{multirow}
\usepackage{array}
\usepackage{breqn}
\usepackage{mathtools}
\usepackage{amsmath}
\usepackage{tikz}
\usepackage[english]{babel}
\usepackage{algorithm}
\usepackage{algorithmic}
\usepackage{graphicx}
\usepackage{tabularray}
\usepackage{tikz}
\usetikzlibrary{positioning, shapes, fit}

\labelformat{subfigure}{\thefigure(#1)}
\newcolumntype{P}[1]{>{\centering\arraybackslash}p{#1}}
\AtBeginDocument{%
  }

\copyrightyear{2025}
\acmYear{2025}
\setcopyright{acmlicensed}\acmConference[ACM MOBICOM '25]{The 31st Annual
International Conference on Mobile Computing and Networking}{November 4--8,
2025}{Hong Kong, China}
\acmBooktitle{The 31st Annual International Conference on Mobile Computing and
Networking (ACM MOBICOM '25), November 4--8, 2025, Hong Kong, China}
\acmDOI{10.1145/3680207.3765261}
\acmISBN{979-8-4007-1129-9/2025/11}

\begin{document}

\title{Experience Paper: Adopting Activity Recognition in On-demand
Food Delivery Business}


\author[Huatao Xu, Yan Zhang, Wei Gao, Guobin Shen, Mo Li]{Huatao Xu$^{1*}$, Yan Zhang$^{2*}$, Wei Gao$^{2}$, Guobin Shen$^{3}$, Mo Li$^{1}$}
\affiliation{%
  \institution{$^1$Hong Kong University of Science and Technology, $^2$Rajax Network Technology (ele.me), \\ $^3$Hong Kong University of Science and Technology (Guangzhou)
  }
  \country{Email:\{huatao, lim\}@ust.hk, yanzhang0@outlook.com, gw84ster@gmail.com, guobinshen@hkust-gz.edu.cn}
}

\def \authors{Huatao Xu, Yan Zhang, Wei Gao, Guobin Shen, Mo Li}
\thanks{Huatao Xu and Yan Zhang contributed equally to the paper. Mo Li is the corresponding author.}


\begin{abstract}
This paper presents the first nationwide deployment of human activity recognition (HAR) technology in the on-demand food delivery industry. We successfully adapted the state-of-the-art LIMU-BERT foundation model to the delivery platform. Spanning three phases over two years, the deployment progresses from a feasibility study in Yangzhou City to nationwide adoption involving 500,000 couriers across 367 cities in China. The adoption enables a series of downstream applications, and large-scale tests demonstrate its significant operational and economic benefits, showcasing the transformative potential of HAR technology in real-world applications. Additionally, we share lessons learned from this deployment and open-source our LIMU-BERT pretrained with millions of hours of sensor data.
\end{abstract}




\maketitle

\section{Introduction}

The rapid advancement of Inertial Measurement Units (IMUs) has revolutionized various domains, ranging from robotics and automotive systems to consumer electronics such as smartphones, smartwatches, and earphones. These compact sensors serve as a cornerstone for a wide array of ubiquitous applications \cite{liu2019real,zhou2019limbmotion,qin2019learning,xu2020touchpass,sun2021idol,xu2021limu}, including user authentication \cite{qin2019learning, xu2020touchpass} and motion tracking \cite{jiang2018ptrack, sun2021idol}. Among these applications, Human Activity Recognition (HAR) has emerged as a pivotal technology that garnered significant attention in research \cite{yao2017deepsense,jiang2015human,saeed2019multi,liu2020giobalfusion, xu2021limu,xu2023practically}. 
\begin{figure}[t!]
  \centering
  \includegraphics[width=0.73\linewidth]{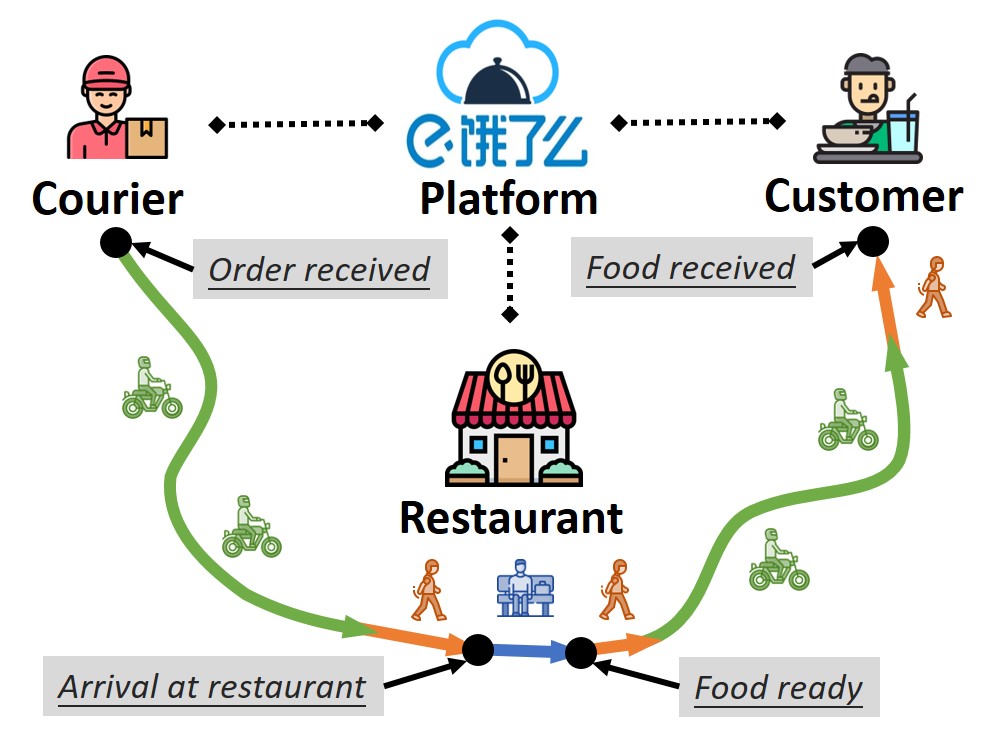}
  \caption{A typical food delivery servicing cycle.}
  \label{fig:scenario_new}
\end{figure}

However, there is no documented experience in the commercial adoption of this technology in large-scale scenarios. This paper addresses this gap by presenting our nationwide experience in applying the activity recognition model to the on-demand food delivery industry \cite{li2020review}, which to our best knowledge, is the first of its kind. Figure \ref{fig:scenario_new} illustrates a typical delivery cycle where the courier rides an electric scooter to reach the restaurant after receiving the order from the platform. Upon arrival, they may walk into the restaurant to pick up the food. If the food is not ready, they may remain stationary and wait until it is ready for collection. The courier then rides the scooter and finally walks to the customer’s location to complete the delivery. Accurately identifying time points, such as the arrival at the restaurant in Figure \ref{fig:scenario_new}, is crucial for the online platform to optimize its decisions. Activity status, on the other hand, provides strong clues for detecting these time points and tracking the courier's progress.

Therefore, we collaborated with Ele.me \cite{fengniao}, the second-largest on-demand food delivery platform in China that operates in over 300 cities and employs a fleet of around half a million couriers, to design a large-scale activity recognition solution. Building upon and extending the original design of a pioneering sensor foundation model LIMU-BERT reported in \cite{xu2021limu}, we leverage IMU sensors embedded in smartphones to recognize the activity and movement status of couriers, such as walking or riding a scooter. By gaining accurate and real-time insights into the status of couriers, the food delivery platform can make more optimized decisions, including price strategy and navigation recommendation, which ultimately enhances courier delivery efficiency and archives significant cost savings for the platform.

Our deployment and evaluation of HAR models were conducted in three phases. Phase I (January 2022–June 2022): A feasibility study was conducted to finalize the design of LIMU-BERT for on-demand food delivery services. This phase included city-scale experimentation in Yangzhou City to evaluate the model's performance and refine its design. Phase II (July 2022–December 2022): LIMU-BERT was pretrained using billions of unlabeled data samples collected from nationwide couriers' smartphones, involving 60K couriers and 1.1K phone models. To evaluate the trained model at scale, we generate coarse-grained activity labels using indoor/outdoor detection and GPS speed data. The evaluation was conducted on a large-scale dataset involving 500,000 couriers, demonstrating that our model achieves over 90\% recognition accuracy in a nationwide deployment. Phase III (June 2023–present): application models supported by LIMU-BERT have been gradually deployed to Ele.me couriers across 367 cities, covering more than 500K couriers. The models currently execute approximately 7.5 billion predictions daily, showcasing their applicability in real-world operations. 

We demonstrate several practical applications that benefited from the adoption of HAR techniques, e.g., trajectory segmentation based on activity status and elevation change detection (identifying whether a courier exhibits significant vertical movement). To evaluate the business impact of the adoption in the on-demand food delivery industry, we conducted comprehensive end-to-end A/B testing using over 1 million orders. The results indicate that adopting HAR technologies reduces the mean absolute error (MAE) of the Estimated Time of Stop (ETS) by 1.8 seconds per order. Additionally, the implementation of a new pricing strategy enabled by LIMU-BERT saves an average of 0.06 yuan per delivery. This translates to an estimated annual savings of approximately 0.44 billion RMB, underscoring the substantial economic benefits of integrating HAR into large-scale operations.

The contributions of this paper are summarized below.
\begin{itemize} 
    \item We showcase the practical value of leveraging human activity recognition technology in real-world commercial applications. 
    \item We share experiences and lessons learned in the training, evaluating, and deploying deep learning-based HAR models for large-scale scenarios. 
    \item We will open-source our LIMU-BERT \footnote{\href{https://github.com/WANDS-HKUST/LIMU-BERT_Experience}{https://github.com/WANDS-HKUST/LIMU-BERT_Experience}}, pretrained with approximately 1.43 million hours of sensor data from 60K subjects and 1.1K phone models, to support further studies and advancements in the field. 
\end{itemize}

Section 2 introduces the foundational concepts of human activity recognition, the on-demand food delivery service, and the design of our HAR model. Section 3 details the deployment process. Section 4 introduces the downstream applications enabled by HAR and the business benefits achieved. Section 5 outlines the lessons learned from the deployment and Section 6 makes a discussion. Section 7 reviews related work, and Section 8 concludes this paper.

\section{Activity Recognition For On-Demand Food Delivery Service}
\subsection{On-demand Food Delivery}
On-demand delivery is a rapidly growing industry worldwide, where couriers transport meals and groceries directly from merchants to customers. To enhance delivery efficiency, platforms require intelligent order dispatch strategies that can optimally match couriers with food orders in real-time. In this paper, we investigate how activity statuses, such as walking or vertical movements, can further support and improve delivery services.

Accurately identifying the time points as shown in Figure \ref{fig:scenario_new} is crucial for optimizing platform decisions. However, the food readiness or pickup times reported by couriers are often subject to objective bias \cite{ding2021nationwide}. As illustrated in Figure \ref{fig:scenario_new}, the courier's activity context provides a natural segmentation of their trajectory, aiding in pinpointing these key time points within specific time ranges. This observation highlights the potential of activity recognition to bring significant value to on-demand food delivery services. We explore other valuable applications that can benefit from accurate activity contexts in Section \ref{sec:downstream}.

\subsection{Challenges}
\begin{figure}[t!]
  \centering
  \includegraphics[width=1.0\linewidth]{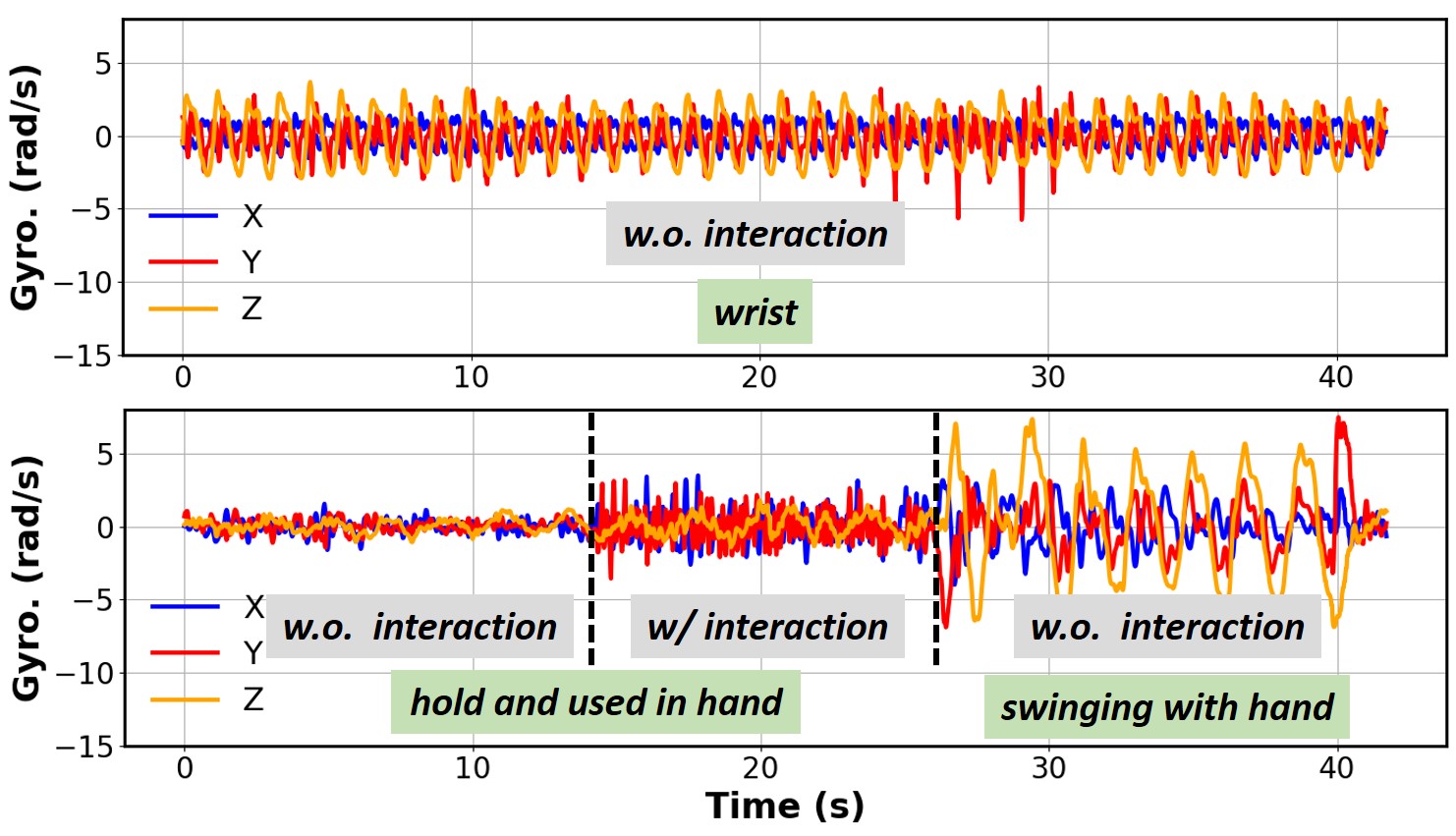}
  \caption{Comparison between gyroscope data in controlled condition (upper part) \cite{shoaib2014fusion} and real condition (lower part). Both two samples are collected when the user is walking. The text in the green boxes shows the placement of smartphones.}
  \label{fig:imu}
\end{figure}
The primary challenge in adopting activity recognition for on-demand food delivery lies in the absence of a comprehensive and large-scale labeled sensor dataset suitable for training and evaluating recognition models. In the ele.me application scenario, there are over 500k couriers across 367 cities in China, using more than 1,000 different phone models. This diversity introduces significant data heterogeneity \cite{stisen2015smart, xu2023practically}, caused by variations in usage patterns, devices, and environmental factors. Collecting labeled data to encompass most scenarios at this scale incurs an immense overhead.

While numerous public IMU datasets exist \cite{stisen2015smart, reyes2016transition, malekzadeh2019mobile, shoaib2014fusion, sztyler2016body, ouyang2021clusterfl} for activity recognition, none, to the best of our knowledge, address such a level of diversity required for large-scale applications. Moreover, most of these datasets were collected under controlled conditions, with smart devices fixed at specific positions, e.g., on the torso. In contrast, real-world scenarios reveal that smartphone placement frequently varies, and users interact with their devices during use. This is especially true for couriers, who regularly handle their smartphones to operate delivery apps, like checking order statuses or contacting customers. Figure \ref{fig:imu} illustrates this disparity by comparing IMU data from a courier with the Shoaib open dataset \cite{shoaib2014fusion}. The controlled dataset, where smartphones are fixed, exhibits stable and regular data patterns. In contrast, real-world data demonstrate large distributional changes due to variations in smartphone placement, and user interactions.
\begin{table}[t!]
    \centering
    \caption{Performance of LIMU-GRU HAR model.}
    \begin{tabular}{cccc}
    \toprule
        Training Dataset & Test Dataset     & Accuracy     & F1-score      \\ \midrule
        Shoaib & Shoaib  & 89.5\% & 89.2\% \\ 
        Shoaib & Yangzhou & 28.2\% & 20.8\% \\ 
        \bottomrule
    \end{tabular}
    \label{tab:dataset:gap}
\end{table}

To further examine the impact of data discrepancies on model performance, we train a LIMU-GRU model, as described in \cite{xu2021limu} using the Shoaib dataset \cite{shoaib2014fusion}, and tested it on the Yangzhou dataset, which is collected under uncontrolled conditions and detailed in Section \ref{sec:phase1}. The Shoaib dataset is selected because it is widely used in existing studies \cite{xu2021limu,xu2023practically,saeed2019multi} and includes a variety of users, activities, and device placements. Table \ref{tab:dataset:gap} illustrates the performance of the model when transferred from the Shoaib dataset to our dataset. While the model achieves high accuracy and F1 scores on the Shoaib dataset, its performance degrades sharply to 0.2 on our dataset. This highlights that data discrepancies can severely hinder the model’s ability to generalize to real-world scenarios.

In addition, the HAR model is expected to be deployed on smartphones for real-time activity recognition. Therefore, another challenge is to ensure the model is lightweight and affordable for most modern phone models. This ensures minimal impact on APK size and energy consumption, which are essential metrics for commercial use.


\subsection{Activity Recognition Design}
\begin{figure}[t!]
  \centering
  \includegraphics[width=1.0\linewidth]{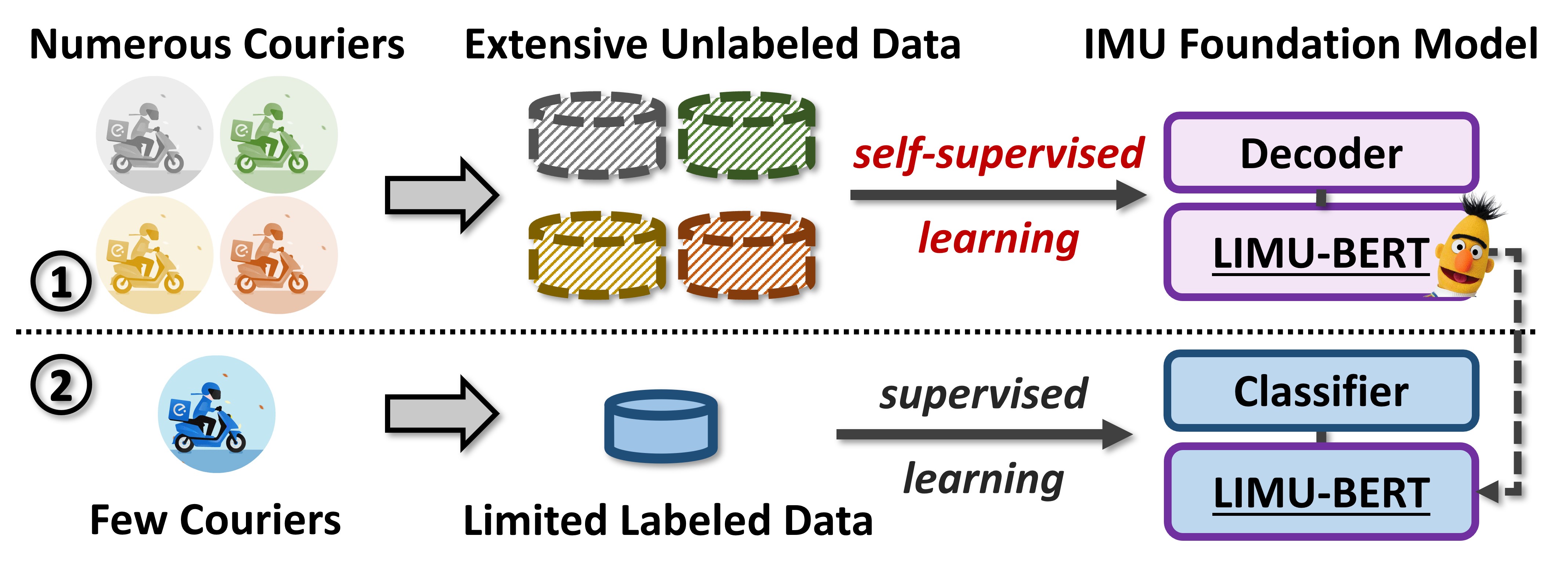}
  \caption{The learning workflow of LIMU-BERT \cite{xu2021limu} in on-demand food delivery scenario.}
  \label{fig:limu}
\end{figure}
\begin{table*}[ht!]
\caption{Overview of three-phase deployment of LIMU-BERT for activity recognition.}
\centering
\scalebox{0.92}{
\begin{tabular}{c|c|c|c|c}
\toprule
\multicolumn{2}{c|}{Phase} & \begin{tabular}[c]{@{}c@{}}\textbf{Phase I}\\ Small-scale evaluation\end{tabular} & \begin{tabular}[c]{@{}c@{}}\textbf{Phase II}\\ Large-sclae evaluation\end{tabular} & \begin{tabular}[c]{@{}c@{}}\textbf{Phase III}\\ Online deployment\end{tabular} \\ \midrule
\multicolumn{2}{c|}{Timeline} & 2022/01-2022/06 & 2022/07-2022/12 & 2023/01-present \\\midrule
\multicolumn{2}{c|}{Area} & sub-city & 367 cities & 367 cities \\ \midrule
\multirow{2}{*}{\begin{tabular}[c]{@{}c@{}}\vspace{-2mm}\\Training \\ Data\end{tabular}} & Labeled & \begin{tabular}[c]{@{}c@{}}10 couriers, 8 phone models\\ 822K samples\end{tabular} & \begin{tabular}[c]{@{}c@{}}10 couriers, 8 phone models\\ 902K samples\end{tabular} & - \\ \cmidrule{2-5}
 & Unlabeled & - & \begin{tabular}[c]{@{}c@{}}60K couriers, 1.1K phone models\\ 858M samples\end{tabular} & - \\ \midrule
\multicolumn{2}{c|}{Evaluation} & \begin{tabular}[c]{@{}c@{}}10 couriers, 8 phone models\\ 97k samples\end{tabular} & \begin{tabular}[c]{@{}c@{}}500K couriers, 1.9K phone models\\ 290M samples\end{tabular} & - \\ \midrule
\multicolumn{2}{c|}{Deployment} & - & \begin{tabular}[c]{@{}c@{}}5 couriers, 4 phone models\\ 3K samples\end{tabular} & \begin{tabular}[c]{@{}c@{}}500K couriers, 1.9K phone models\\ {7.5B samples per day}\end{tabular}\\ \midrule
\multicolumn{2}{c|}{Performance} & \begin{tabular}[c]{@{}c@{}}(3-activity classification)\\ 89.2\% accuracy\\ 88.1\% F1-score\end{tabular} & \begin{tabular}[c]{@{}c@{}}(riding and non-riding classification)\\ 90.1\% precision\\ 94.5\% recall\end{tabular} & - \\
\bottomrule
\end{tabular}
}
\label{tab:deployment}
\end{table*}

Despite the significant challenges posed by large-scale HAR, we identify a key opportunity within the ele.me application scenario: extensive sensor data can be readily collected through delivery apps. While these data are unlabeled, they can still be utilized to enhance model performance using self-supervised learning techniques \cite{jaiswal2020survey, liu2021self,haresamudram2022assessing,logacjov2024self}. To capitalize on this, we adopt LIMU-BERT \cite{xu2021limu}, a state-of-the-art IMU foundation model, which effectively harnesses the potential of large-scale unlabeled data and thus well matches our context.

Figure \ref{fig:limu} illustrates the core concept of LIMU-BERT applied to the on-demand food delivery scenario, comprising two distinct learning phases. In the first phase, LIMU-BERT is pre-trained using extensive unlabeled data. During training, we randomly mask a subset of the IMU sequence data (e.g., 20 out of 120 samples) and jointly train LIMU-BERT along with a decoder to reconstruct the partially masked inputs, enabling the model to learn high-level representations from sensor data. In the second phase, the pre-trained LIMU-BERT is reused and connected to a classifier model. Together, they are fine-tuned using a limited amount of labeled data for activity recognition tasks. After the supervised learning phase, the combined LIMU-BERT and classifier model is deployed on smartphones, where it performs real-time activity inference for couriers.

We believe LIMU-BERT is well-suited for our scenario because it can effectively leverage unlabeled data and achieves high performance even when labeled data is scarce \cite{xu2021limu}. Furthermore, the original model described in \cite{xu2021limu} consists of only 71K parameters, making it lightweight and capable of supporting real-time activity recognition on smartphones.

\section{Deployment and experience}
\begin{figure*}[t!]
\begin{subfigure}{.24\linewidth}
  \centering
  \includegraphics[width=1.0\linewidth]{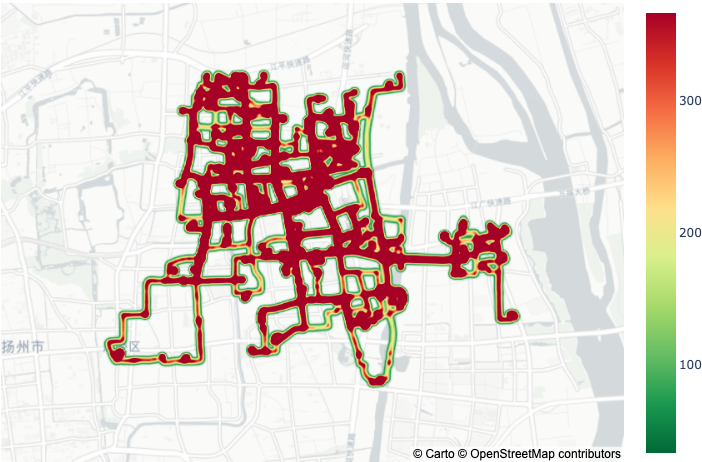}
  \caption{Labeled data.}
  \label{fig:dis:label}
\end{subfigure}
\begin{subfigure}{.24\linewidth}
  \centering
  \includegraphics[width=1.0\linewidth]{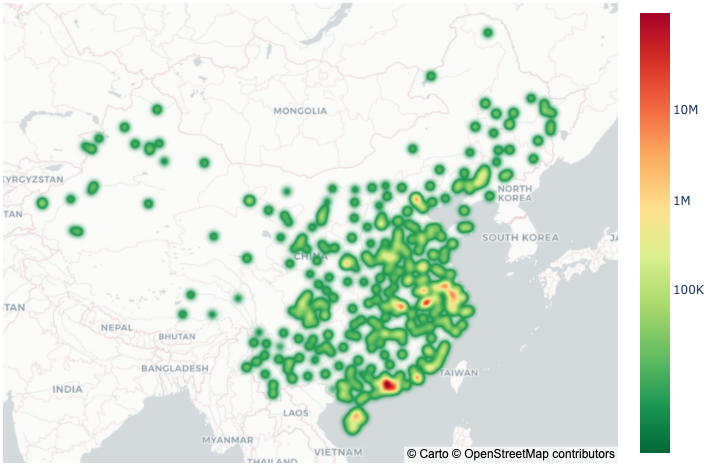}
  \caption{Unlabeled data.}
  \label{fig:dis:unlabel}
\end{subfigure}
\begin{subfigure}{.24\linewidth}
  \centering
  \includegraphics[width=1.0\linewidth]{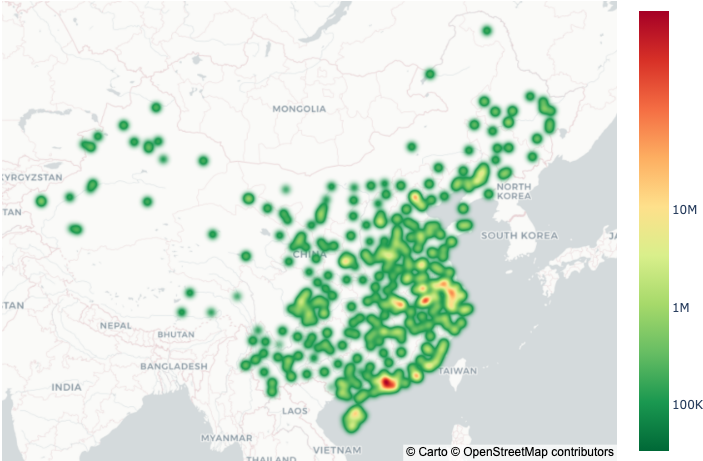}
  \caption{Inference data.}
  \label{fig:dis:inference}
\end{subfigure}
\begin{subfigure}{.24\linewidth}
  \centering
  \includegraphics[width=1.0\linewidth]{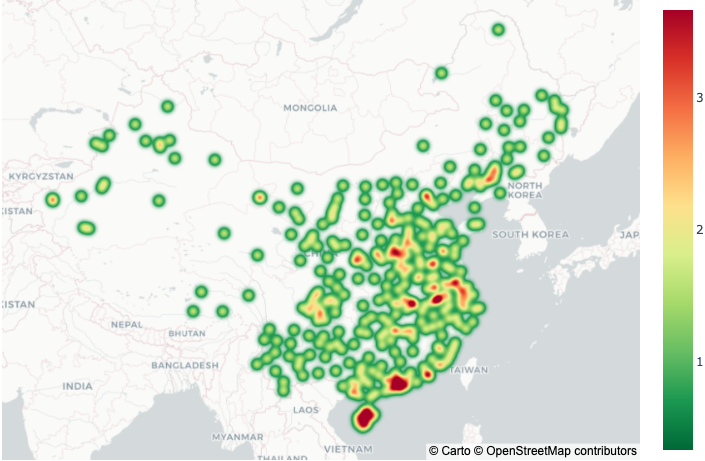}
  \caption{City density.}
  \label{fig:dis:city}
\end{subfigure}
\caption{The spatial distribution of labeled, unlabeled, and inference data for the deployment. (d) displays the city density (the number of cities located within a defined area) of the deployment.}
\label{fig:distribution}
\end{figure*}

Since January 2022, we have implemented the deployment of HAR models in three consecutive phases: the small-scale evaluation phase (Phase I, Jan.–Jun. 2022), the large-scale evaluation phase (Phase II, Jul.–Dec. 2022), and the online deployment phase (Phase III, Jan. 2023–present). Table \ref{tab:deployment} summarizes key features of the three phases. We will elaborate on their details in the following subsections. Figure \ref{fig:distribution} shows the spatial distribution of the data involved in the deployment.



\subsection{Phase I Small-scale Evaluation}\label{sec:phase1}
\begin{table}[ht!]
    \centering
    \caption{Impact of sampling rate.}
    \begin{tabular}{cccc}
    \toprule
        Dataset  & Sampling Rate   & Accuracy     & F1-score      \\ \midrule
        Shoaib & 20Hz  & 90.0\% & 90.1\% \\
        Shoaib & 10Hz  & 89.5\% & 89.2\% \\ 
        \bottomrule
    \end{tabular}
    \label{tab:sampling}
\end{table}

\textbf{Design customization.} Building on the original LIMU-BERT design \cite{xu2021limu}, we made several adjustments to better suit our scenario. First, we reduced the IMU data sampling rate from 20 Hz to 10 Hz. This adjustment helps lower data collection costs, including data transmission and inference, and also reduces sequence length, thereby decreasing model complexity for the same time window. To evaluate the impact of this change, we conducted a comparative experiment using the Shoaib dataset \cite{shoaib2014fusion}. As shown in Table \ref{tab:sampling}, reducing the sampling rate to 10 Hz resulted in only minor performance degradation compared to 20 Hz. To this end, we adopted a 10 Hz sampling rate for our implementation. Second, we reduced the window size from 120 (equivalent to 12 seconds at 10 Hz) to 60 (equivalent to 6 seconds at 10 Hz) to enable more timely inference. Therefore, each input sample to the model is represented as a matrix in $\mathbb{R}^{60 \times 6}$, where 60 denotes the window size and 6 corresponds to the total number of dimensions from the accelerometer and gyroscope sensors. Third, all the parameters in LIMU-BERT are fine-tuned with the classifier during the second learning phase, as illustrated in Figure \ref{fig:limu}, to enhance the model's fitting capabilities. We define BERT model with $R_{num}=4$, $A_{dim}=4$, $H_{dim}=36$, and $F_{dim}=72$ \cite{xu2021limu}. We adopt a Gated Recurrent Unit (GRU)-based classifier comprising three stacked GRU layers with hidden sizes of 20, 20, and 10, respectively, following the design in \cite{xu2021limu}. During training, both the pretrained BERT model and the GRU classifier are jointly trained for 700 epochs, using a learning rate of 0.001 and a batch size of 128.

\textbf{Labeled dataset collection.} The data differences observed in Figure \ref{fig:imu} prompted us to collect a labeled dataset under uncontrolled real-world conditions. In June 2020, we conducted a data collection effort in Yangzhou City, resulting in the dataset referred to as the \textit{Yangzhou dataset} in this paper. To minimize the impact of the data collection process on couriers' natural behaviors, we employed a video-assisted approach. Couriers were equipped with wearable cameras mounted on their chests to record videos during their daily work. The video footage (about 420 hours in total) was carefully analyzed, and a third-party company was engaged to annotate the couriers' activity statuses. The collected sensors are the accelerometer and gyroscope sampled at 10 Hz. As shown in Figure \ref{fig:scenario_new}, to better align the model with the specific needs and dynamics of the courier industry, we choose a three-class classification task: \textit{still}, \textit{walking}, and \textit{riding} (the latter refers to riding an electric scooter or motorcycle). We merge \textit{running} into the \textit{walking} category as their distinction does not directly affect commercial applications.

Figure \ref{fig:dis:label} shows the spatial distribution of the Yangzhou dataset, which includes data from 10 delivery riders with 8 unique smartphone models (including Xiaomi, Huawei, and iPhone) for 7 days, with each rider contributing an average of 13,000 samples per day. The number of labeled samples for the activities \textit{still}, \textit{walking}, and \textit{riding} are 277,140, 168,007, and 456,629, respectively, resulting in a total of 901,776 samples. However, the Yangzhou dataset covers minor aspects of users, devices, and environments, which is still limited compared with sensor data collected from nationwide couriers. The distribution of this dataset may differ from other couriers due to the potential Hawthorne effect [31], where couriers may alter their behavior, for example, becoming more likely to follow safety guidelines when they know they are being observed \cite{sedgwick2015understanding}. 

\begin{table}[ht!]
    \centering
    \caption{Performance comparison of different models on the Yangzhou dataset.}
    \begin{tabular}{ccc}
    \toprule
        Method    & Accuracy     & F1-score      \\ \midrule
        FFT+LR  & 81.4\% & 80.1\% \\
        DCNN  & 80.6\% & 79.7\% \\
        LIMU-BERT  & 89.2\% & 88.1\% \\ 
        \bottomrule
    \end{tabular}
    \label{tab:eva:bench}
\end{table}
\textbf{Model training and evaluation.} We first evaluate the performance of LIMU-BERT (w.o. pretraining) using the Yangzhou dataset by comparing it with two alternative approaches: (1) FFT + Logistic Regression: We apply the Fast Fourier Transform (FFT) to the sensor data and use a logistic regression model for activity classification. (2) DCNN \cite{yang2015deep}: We implement a Convolutional Neural Network (CNN)-based model for activity recognition. All models were trained using the same and 90\% labeled samples and tested on the 10\% of samples, and their performance is summarized in Figure \ref{tab:eva:bench}. The FFT + Logistic Regression model achieves an accuracy of 0.81 and an F1-score of 0.80. The DCNN model shows slightly lower results, with an accuracy of 0.80 and an F1-score of 0.79. Both approaches perform significantly worse than LIMU-BERT, which demonstrates superior accuracy and a high F1 score, indicating balanced performance across all activity classes. These findings suggest that traditional models, such as FFT with logistic regression and DCNN, struggle to effectively capture the complexities and variability present in real-world datasets. In contrast, transformer-based models like LIMU-BERT excel at understanding sequential dependencies and capturing intricate patterns, resulting in markedly improved performance in predicting courier activities. However, the data collected from nationwide couriers is significantly more complex than the Yangzhou dataset, meaning the performance of LIMU-BERT on the Yangzhou dataset may not fully reflect its effectiveness in a nationwide deployment.

We also experiment with pretraining the LIMU-BERT model using an external unlabeled dataset, as depicted in Figure \ref{fig:limu} phase one. Specifically, we use the Shoaib dataset \cite{shoaib2014fusion} for pretraining and then fine-tune the model using the labeled data from the Yangzhou dataset. However, the model's performance degrades significantly, with the accuracy and F1-score dropping to 75.9\% and 75.6\%, respectively. This degradation is likely due to the substantial distribution gap between the two datasets, as highlighted in Table \ref{tab:dataset:gap}. These findings further emphasize the importance of collecting a large-scale unlabeled dataset tailored to our application scenario.

\subsection{Phase II: Large-scale Evaluation}
\begin{table*}[t!]
    \centering
    \caption{Comparison of LIMU-BERT with or without pretraining across different percentages of training labels.}
    \label{tab:performance:label}
    \begin{tabular}{ccccccccccc}
        \toprule
        Labeling rate& \multicolumn{2}{c}{0.10\%} & \multicolumn{2}{c}{1\%} & \multicolumn{2}{c}{10\%} & \multicolumn{2}{c}{20\%} & \multicolumn{2}{c}{90\%} \\ \cmidrule(r){1-1}
        \cmidrule(r){2-3} \cmidrule(r){4-5} \cmidrule(r){6-7} \cmidrule(r){8-9} \cmidrule(r){10-11}
        Metric & Acc & F1 & Acc & F1 & Acc & F1 & Acc & F1 & Acc & F1 \\
        \midrule
        LIMU-BERT w/ pretraining & 70.10\% & 66.50\% & 81.20\% & 80.20\% & 87.60\% & 86.20\% & 88.90\% & 87.8\% & 90.50\% & 89.60\% \\
        LIMU-BERT w/o pretraining & 64.50\% & 60.90\% & 76.90\% & 75.80\% & 84.40\% & 83.10\% & 86.6\% & 85.40\% & 89.20\% & 88.10\% \\
        \bottomrule
    \end{tabular}
\end{table*}

\textbf{Unlabeled dataset collection}. 
To pre-train LIMU-BERT, we randomly sample an unlabeled dataset from sensor data collected from nationwide couriers. This dataset comprises a total of 847,684,084 samples, encompassing data from 1.1K smartphone models and 60K couriers across China. As summarized in Table \ref{tab:deployment}, the unlabeled dataset is significantly more comprehensive than the Yangzhou dataset and Figure \ref{fig:dis:unlabel} shows the spatial distribution of the unlabeled dataset. 


\textbf{Model re-training and evaluation}.
We pre-train LIMU-BERT on four Tesla-V100-32G with a batch size of 128 across 8,000 epochs. The training loss is mean square errors (MSE) between the original data and reconstructed data from LIMU-BERT. We adopt the Adam optimizer with a base learning rate of 0.001. The whole training process takes about 780 GPU hours. Then, we fine-tune the model using labeled data from the Yangzhou dataset to construct an HAR.

\begin{figure}[t!]
\begin{subfigure}{.46\linewidth}
  \centering
  \includegraphics[width=\linewidth]{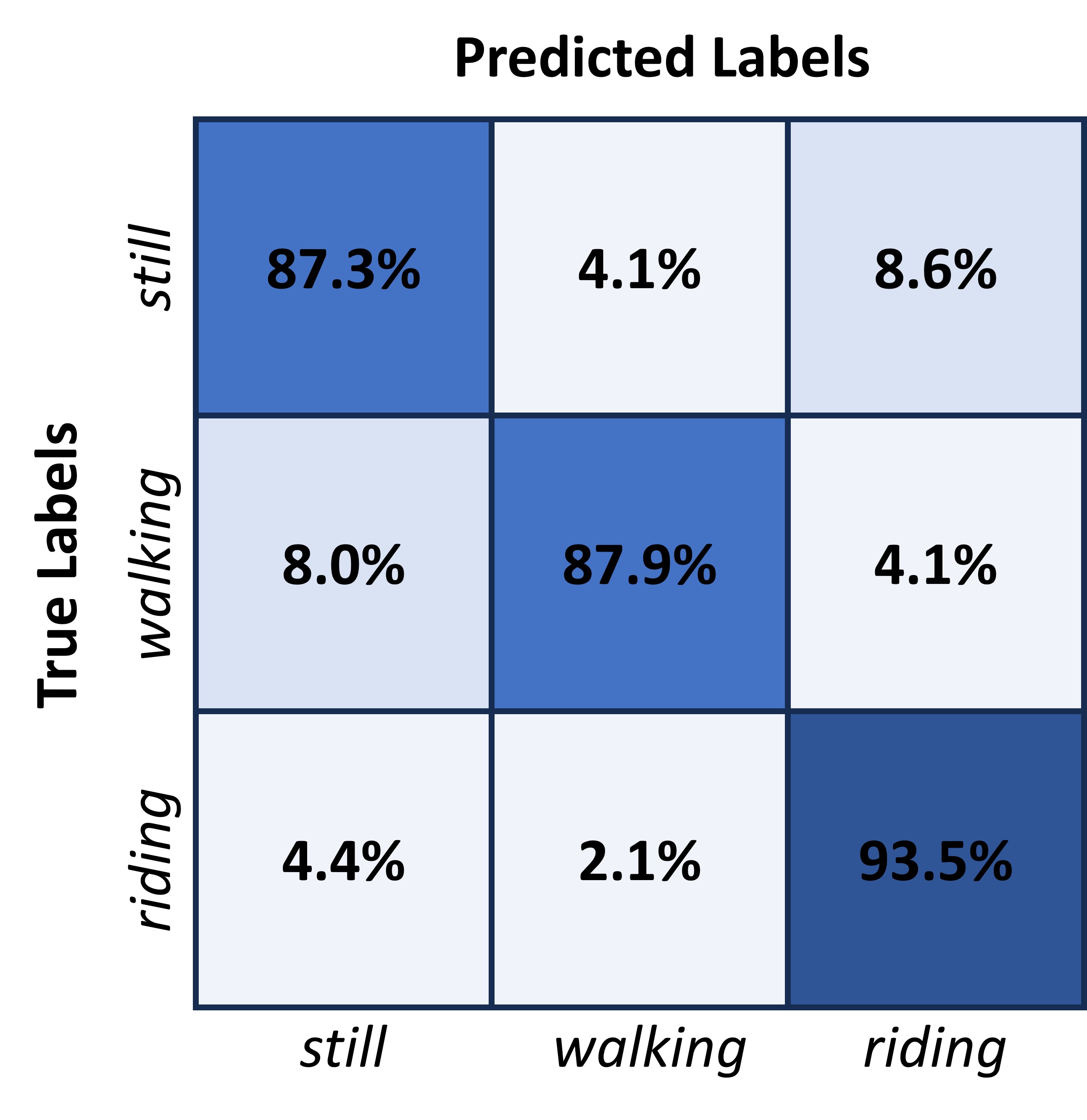}
  \caption{Yangzhou dataset.}
  \label{fig:cm:1}
  \end{subfigure}
 \hspace{1em}
 \begin{subfigure}{.46\linewidth}
  \centering
  \includegraphics[width=0.9\linewidth]{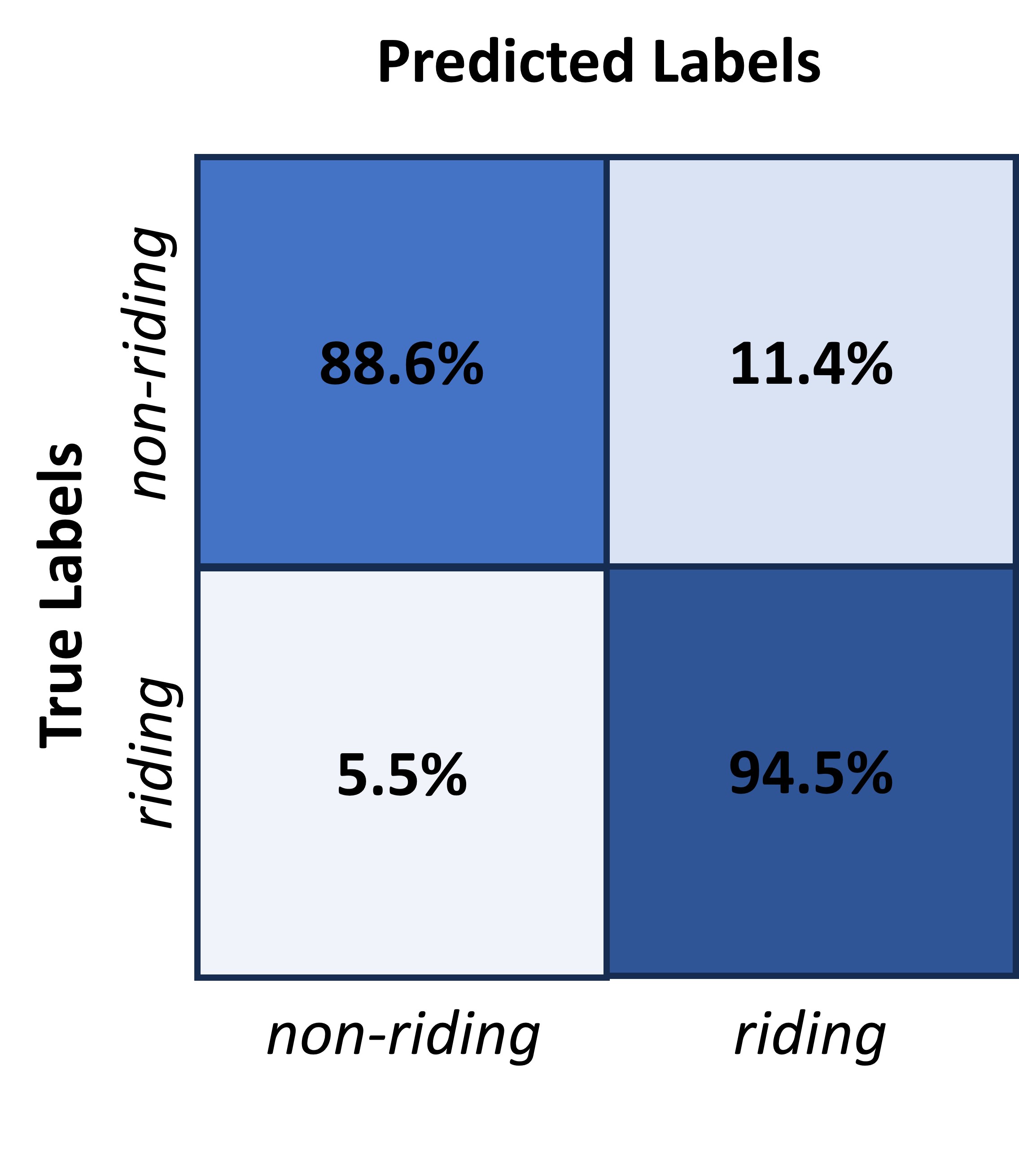}
  \caption{Rule-based dataset.}
  \label{fig:cm:2}
\end{subfigure}
\caption{Confusion matrices illustrating the evaluation results on the second phase. The two results are evaluated on 97K and 290M samples, respectively.}
\label{fig:cm}
\end{figure}

Compared to the model used in Phase I, the performance of new model on the Yangzhou dataset improves slightly, achieving 90.5\% accuracy and 89.6\% F1-score. Figure \ref{fig:cm:1} illustrates the detailed confusion matrix and most activities are accurately recognized. To further validate our approach, we simulated large-scale deployment conditions, where most data are unlabeled. We experimented with varying ratios of labeled samples from the Yangzhou dataset to train the recognition model. This approach enabled us to evaluate the LIMU-BERT's ability to effectively leverage unlabeled data and assess its suitability for widespread adoption. Table \ref{tab:performance:label} presents the performance of LIMU-BERT with and without pretraining using the extensive unlabeled dataset (Phase 1 in Figure \ref{fig:limu}). The model that includes the pretraining stage significantly outperforms the one without it. For example, it achieves a 4.3\% higher accuracy and a 4.4\% higher F1-score when only 1\% of labeled samples are used. More importantly, the pretraining-enabled model consistently delivers better results across all cases, underscoring the necessity of pretraining for effective large-scale deployment.

\textbf{Small-scale deployment.} To further validate our model’s performance, we developed a front-end annotation application and tested it in an online inference scenario. The evaluation involved five couriers using four different phone models, covering approximately 3,000 samples for the three-class classification task. The results showed a slight performance decline, with accuracy decreasing to 88.82\% and the F1 score dropping to 83.07\%. The performance degradation could be attributed to the impact of the annotation process (e.g., smartphone interactions) on the couriers' natural behaviors. Despite this, the results demonstrate the model’s overall effectiveness and highlight its potential, motivating us to conduct a large-scale evaluation.


\textbf{Large-scale evaluation}.
In the second phase of our evaluation, the primary challenge is the lack of large-scale ground-truth activity labels. To address this, we focus on the large-scale assessment of riding activity classification, a critical aspect for the business, and propose a rule-based approach to derive labels from the courier's environment. This method builds on IODetector \cite{zhou2022experience}, which combines multiple sensor inputs for indoor/outdoor detection, and GPS speed data. Specifically, an activity is classified as \textit{riding} if the courier is detected as being outdoors with a GPS speed exceeding 4 m/s. Conversely, any activity recorded indoors is categorized as \textit{non-riding}. To align the motion types, we post-process the model outputs by merging the \textit{still} and \textit{walking} categories into a single non-riding class. The results of this classification task revealed promising accuracy metrics as shown in Figure \ref{fig:cm:2}. For riding activities, LIMU-BERT achieves a precision of 90.1\% and a recall of 94.5\%. 

Overall, the results from our evaluation demonstrate robust performance in the task of riding activity classification. The combination of both video annotation data and the established rule-labeling approach has proven to be effective in delivering high precision and recall metrics. These measures indicate the model's capability to reliably capture and distinguish between riding and non-riding activities across complex environments.

%

\subsection{Phase III: Online Deployment}
\begin{figure*}[ht!]
    \centering
    \begin{minipage}[t]{0.32\textwidth}
        \centering
        \includegraphics[width=0.7\linewidth]{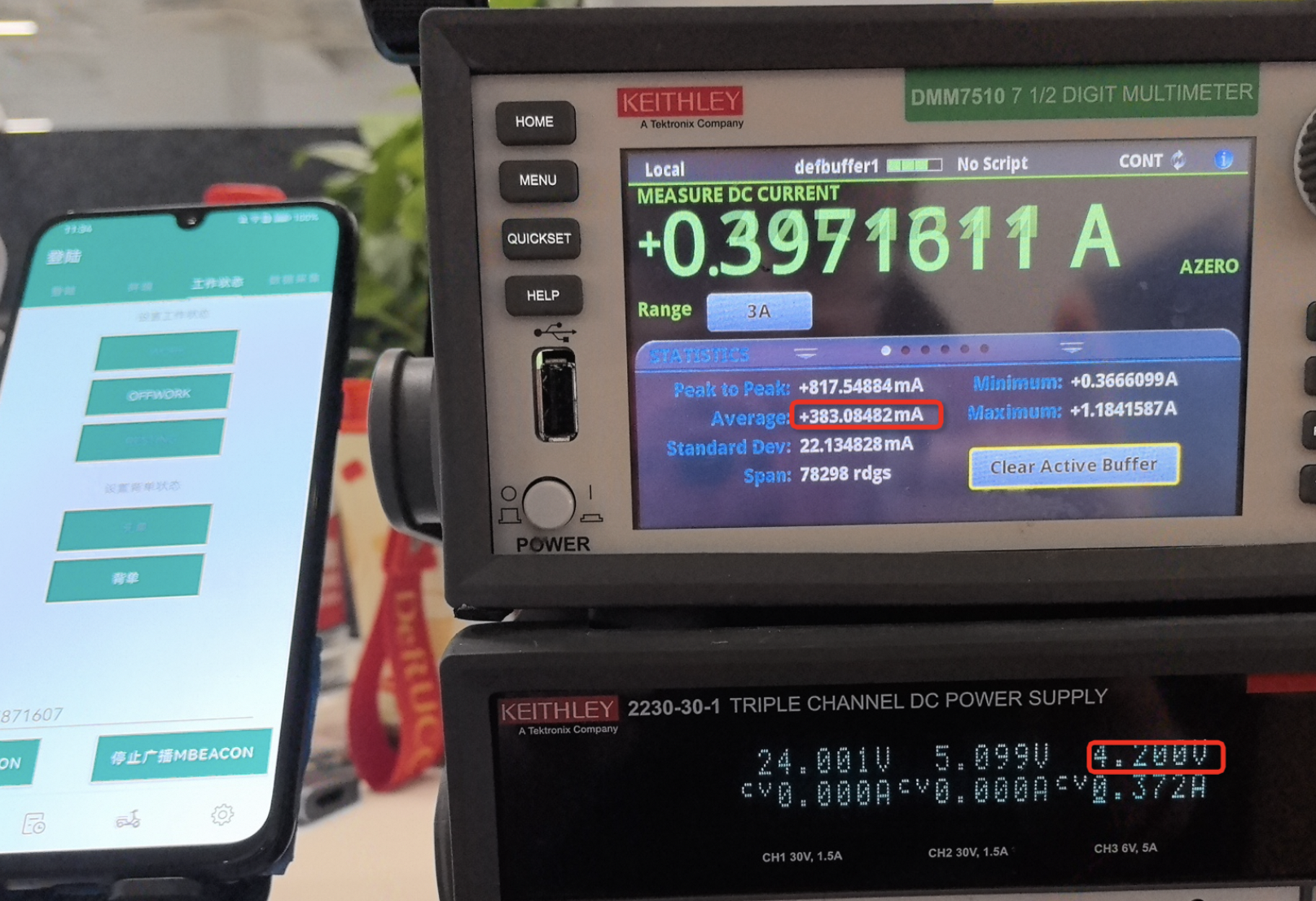}
  \caption{Power measurement.}
  \label{fig:battery}
    \end{minipage}
    \hfill
    \begin{minipage}[t]{0.65\textwidth}
        \centering
        \vspace{-3cm}
        \captionof{table}{Power consumption test results for various SDK module configurations.}
        \label{tab:power}
        \begin{tabular}{|c|c|c|c|}
        \hline
        Active SDK Modules & Voltage (V) & Current (mA) & Power (mW) \\ \hline
        All & 4.2 & 432.05 & 1814.61 \\ \hline
        All excluding edge computing  & 4.2 & 429.05 & 1802.01 \\ \hline
        None & 4.2 & 383.08 & 1608.94 \\ \hline
        \end{tabular}
    \end{minipage}
\end{figure*}
To deploy our recognition model on smartphones, we developed an SDK consisting of three primary modules. 
\begin{itemize}
\item Data collection module: This module provides a listening service for raw sensor data, enabling the acquisition of data from multiple sensors within the Android system. To support evaluation with other sensors and meet the needs of various applications, we also collect data from additional sensors, such as barometers and satellite signals.
\item Edge computing module: This module manages the on/off functionality of the data collection process and facilitates the prediction of couriers' activities using locally collected sensor data. This includes tasks such as loading the model and performing inference.
\item Control module: This module allows dynamic configuration of settings for data collection and model prediction, such as sampling rate. It also supports data compression and uploads for both raw sensor data and prediction results. Additionally, it caches data locally when network conditions are poor, ensuring the data can be uploaded later when connectivity is restored.

\end{itemize}

Initially, we compile the package using PyTorch Lite \cite{paszke2019pytorch}, which however leads to an unacceptable 13 MB increase in app size. It poses a significant issue, as app stores impose size limits for downloads over cellular networks. To address this, we tailor the integration process to minimize the application's size. Upon unpacking the package, we discover that the size increase was primarily due to large dependent libraries, particularly the libpytorch\_jni.so files, which were 41.2 MB for arm64-v8a and 31.7 MB for armeabi-v7a. To resolve this, we manually compile the Android version of PyTorch and implemented a dynamic loading strategy. This approach excludes the large SO libraries from the APK during packaging, allowing them to be fetched from a server at runtime only when needed. As a result, we reduce the package size significantly, with the new package increasing by just 100 KB. Despite this reduction, we retain the ability to load essential libraries on demand, ensuring optimized performance while preserving a seamless experience for couriers.


\textbf{SDK overhead evaluation}. 
To ensure minimal impact on energy consumption, we conducted a series of rigorous tests. Instead of relying on estimations from Android's battery status, we opted for direct measurements using a power monitor after removing the device's battery, as shown in Figure \ref{fig:battery}. This approach allowed for a more precise assessment of the actual power usage, eliminating any discrepancies or inaccuracies inherent to software-based estimations. As shown in Table \ref{tab:power}, our experimental results demonstrated that integrating our SDK increased power consumption by approximately 13\%. A substantial portion of this increase, nearly 10\%, was attributed to the continuous search for satellite signals, especially in indoor environments. Conversely, computational processes related to the SDK contributed less than 1\% to the overall increase. To mitigate this, we implemented strategies to automatically disable non-essential modules and sensors when satellite connectivity is non-critical or when the device is indoors \cite{zhou2012iodetector}. These optimizations help reduce energy consumption, thereby improving the user experience and extending the device's battery life between charges.

\textbf{Launch details}. 
The deployment of LIMU-BERT began in December 2023 and has quickly achieved extensive coverage. Currently, the system supports approximately 500,000 daily active couriers across 367 cities as shown in Figure \ref{fig:dis:inference}, processing around 20 million food delivery orders daily. The diversity of devices is remarkable, with LIMU-BERT operating on about 1,900 distinct phone models. On average, the system performs predictions every two seconds, resulting in an astounding 7.5 billion predictions per day. This expansive reach and high operational frequency highlight the robustness and scalability of LIMU-BERT in real-world scenarios, demonstrating its ability to deliver reliable performance for wide delivery scenarios. To ensure the stability and effectiveness of LIMU-BERT, we implement a robust monitoring system that leverages cloud data transmitted from the SDK. This system primarily tracks the model's coverage in relation to both orders and couriers.




\section{Downstream Applications and Business benefits}\label{sec:downstream}
\begin{figure}[t!]
  \centering
  \includegraphics[width=0.8\linewidth]{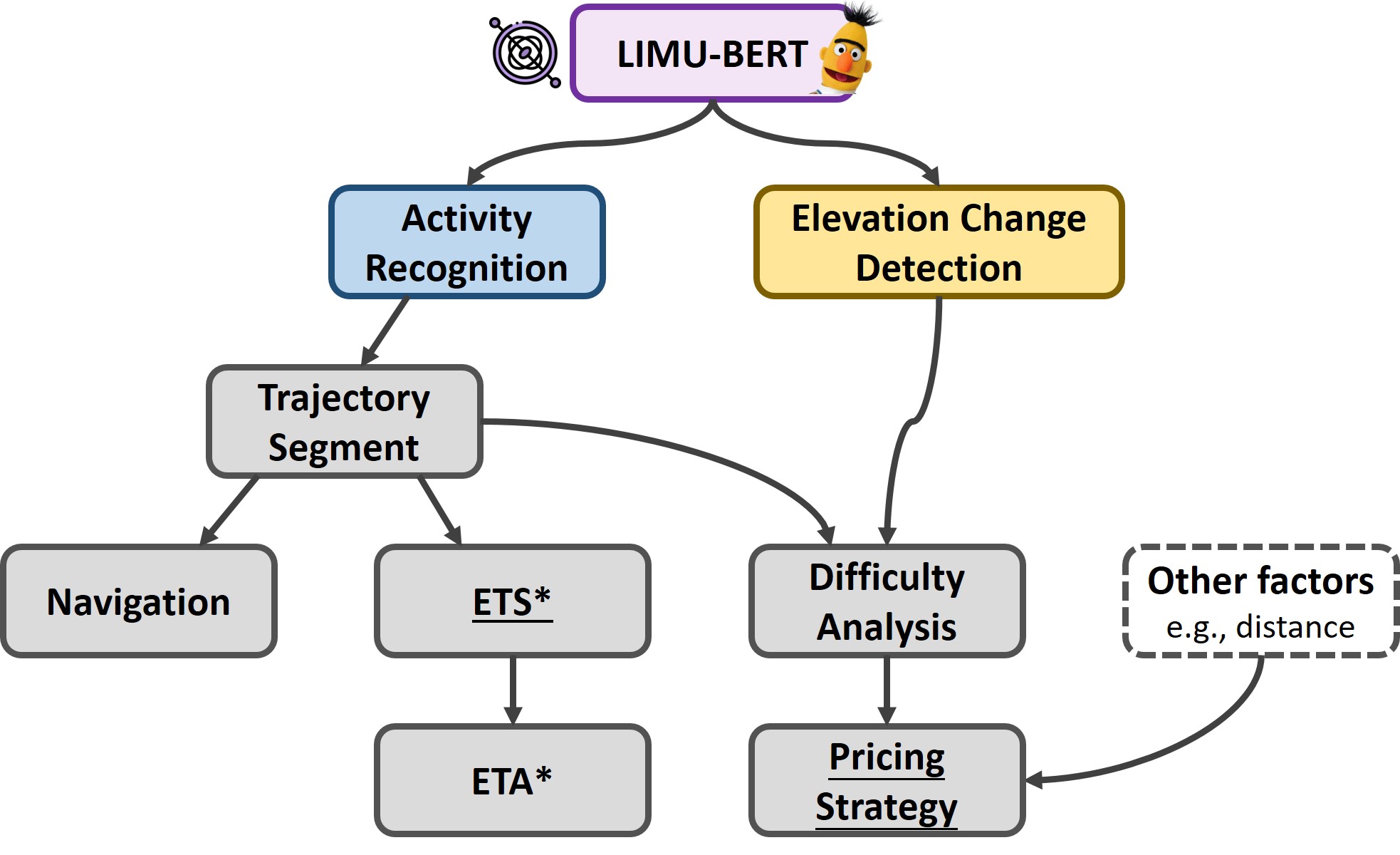}
  \caption{The downstream applications of LIMU-BERT. The ETS and ETA indicate the estimated time of stop and arrival, respectively.}
  \label{fig:downstream}
\end{figure}
The adoption of LIMU-BERT enables several downstream applications as shown in Figure \ref{fig:downstream}. These enhancements optimize delivery routing and timing, allowing the company to efficiently handle a higher volume of orders with the same or fewer resources. This is crucial in today’s fast-paced market, where customer expectations for timely delivery are constantly increasing. The following subsections will provide details of these applications.
\subsection{Trajectory Segmentation and Navigation}
\begin{figure*}[t!]
\begin{subfigure}{.47\linewidth}
  \centering
  \includegraphics[width=1.0\linewidth]{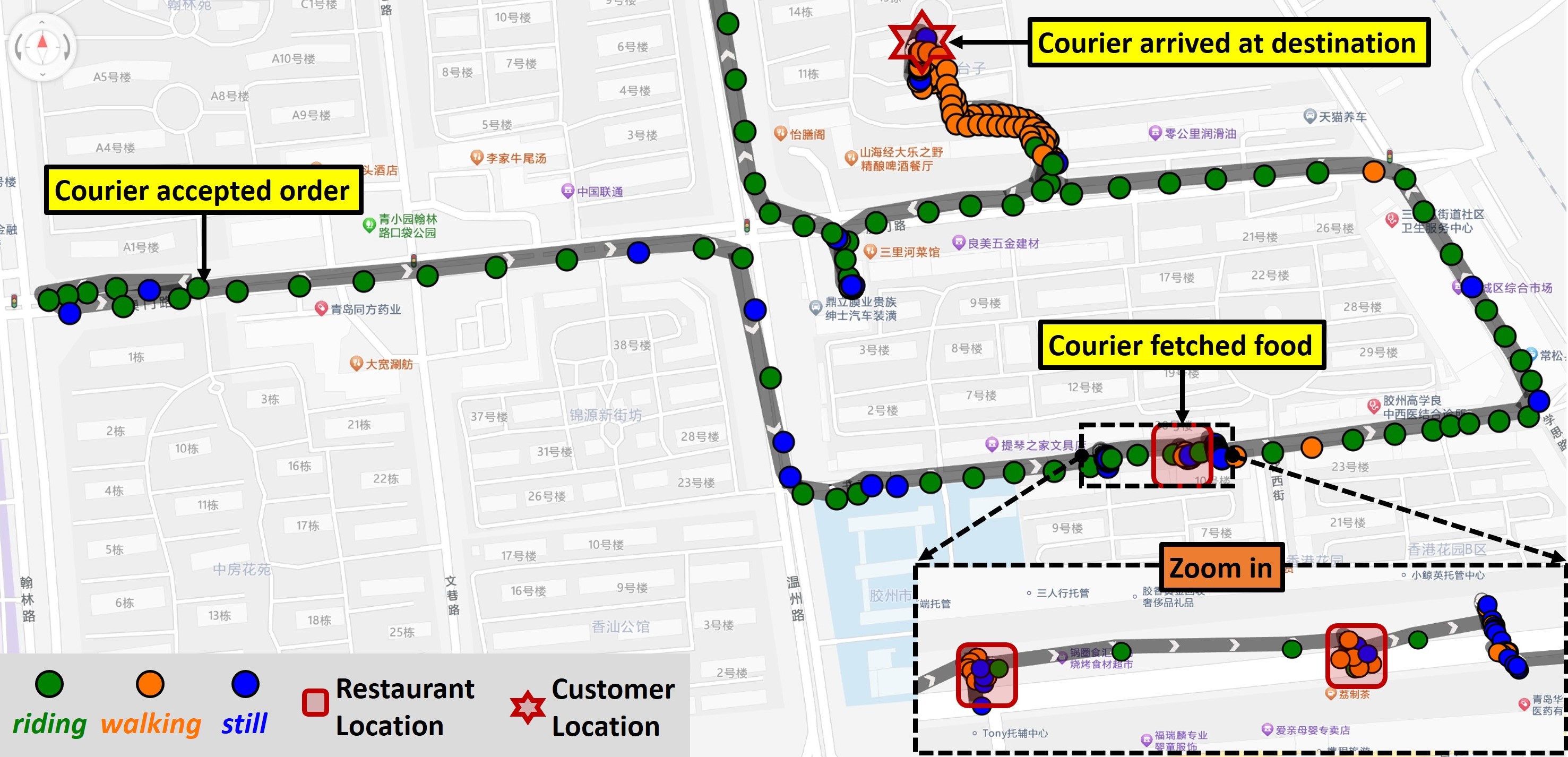}
  \caption{Trajectory with activity status.}
  \label{fig:traj1}
\end{subfigure}
\hspace{1em}
\begin{subfigure}{.47\linewidth}
  \centering
  \includegraphics[width=1.0\linewidth]{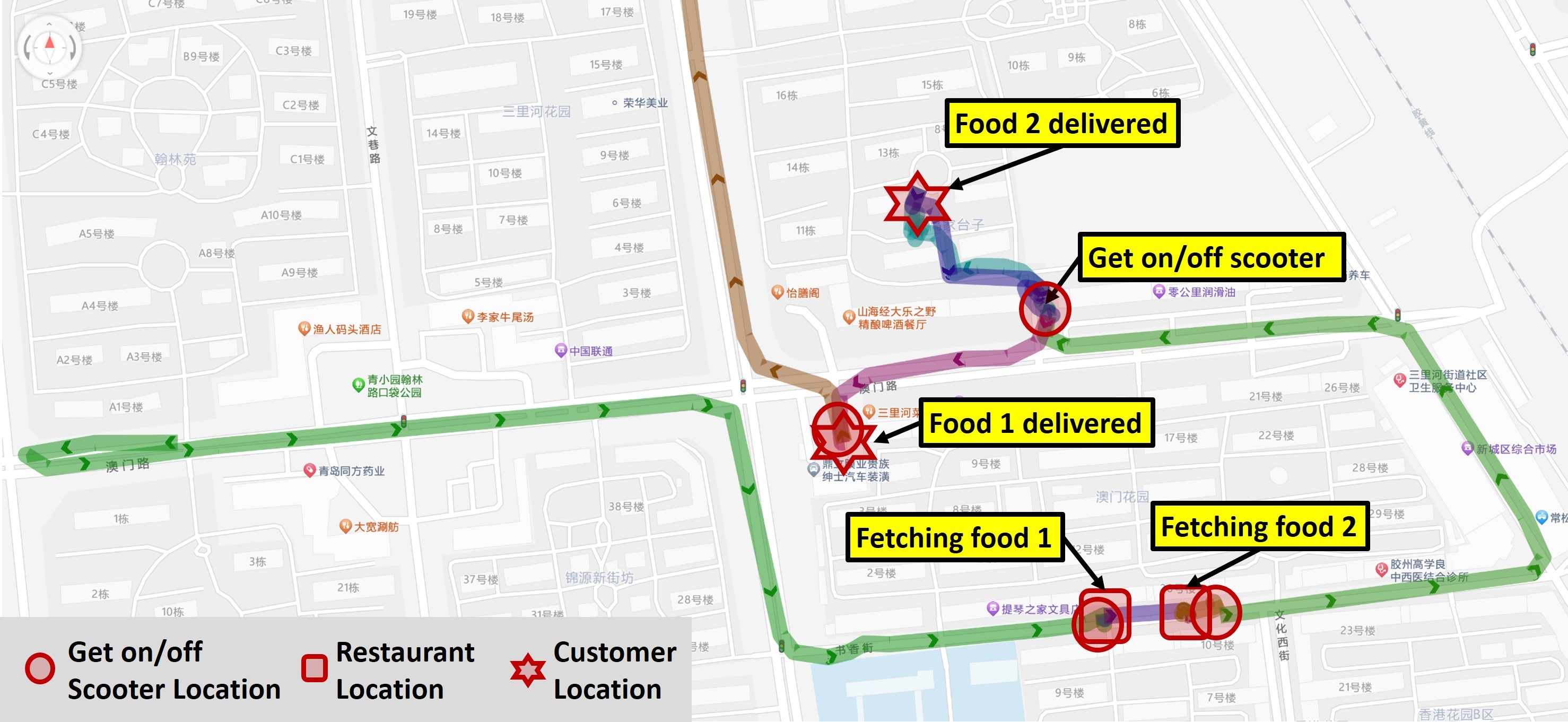}
  \caption{Trajectory segmentation results.}
  \label{fig:traj2}
\end{subfigure}
\caption{Trajectory segmentation application of human activity recognition models.}
\label{fig:traj}
\end{figure*}
As shown in Figure \ref{fig:scenario_new}, the activity status of couriers can assist in identifying key time points during the delivery process, such as their arrival at the restaurant. Figure \ref{fig:traj1} illustrates a trajectory with detected activity statuses and annotated ground-truth time points. In this case, we observe that after accepting an order, the courier rides their scooter to the restaurant to fetch the food. Upon arrival, they stop riding and begin walking or remain stationary. Following this, they ride towards the customer's location and walk to the final destination, as the customer resides in a gated community where scooters are not permitted. After delivering the food, the courier returns to their scooter to begin the next delivery. Notably, when the road is flat and the scooter moves at a constant speed, IMU sensors may not effectively distinguish between riding and being still, which results in occasional false "still" points during the riding process. However, the overall activity statuses are accurate, providing clear insights into key behaviors throughout the delivery process.

Therefore, we employ LIMU-BERT for trajectory segmentation during the delivery process, with the implementation outlined in Algorithm \ref{al:segmentation}. First, the activity labels recognized by LIMU-BERT undergo a smoothing process over fixed time windows to reduce noise and erratic fluctuations. These smoothed activity states are then clustered based on similar activities, such as riding or walking, to identify distinct activity segments. Next, the GPS data corresponding to these clusters is refined to ensure alignment with the identified activity segments. The refined GPS data is then segmented based on the activity clusters, with additional physical constraints applied to ensure plausible transitions over time. For example, locations where couriers get on and off their scooters are constrained to be geographically close to maintain consistency. Figure \ref{fig:traj2} demonstrates the segmentation results for the trajectory shown in Figure \ref{fig:traj1}. Key locations, such as where the courier gets on and off the scooter, are clearly identified. In real scenarios, couriers often handle multiple delivery orders simultaneously to improve efficiency. 

To evaluate the efficacy of our LIMU-BERT-based trajectory segmentation, we implemented a twofold assessment approach. First, we randomly sampled 1,000 delivery orders and performed manual labeling in conjunction with street view images to establish accurate trajectory segments. We achieve a classification accuracy of 95.2\% with this comprehensive labeling process. Second, we specifically examined the precision of the transition points where couriers shift from riding to walking by verifying whether these points fell within a predefined vicinity of their projected locations on main roads. This verification yielded an accuracy of 88.7\% over 20 million orders, demonstrating the algorithm's capability to reliably identify critical transition zones. 

\begin{algorithm}[t!]
    \caption{Trajectory Segmentation with Activity Recognition Integration}
    \label{al:segmentation}
    \begin{algorithmic}[1]
        \STATE \textbf{Step 1: Smooth Activity States}
        \STATE acts $\gets$ [] 
        \FOR{each point in trajectory}
            \STATE acts.append(point.getActState())
        \ENDFOR
        \STATE smoothActs $\gets$ smoothActs(acts, window_size) 
        
        \STATE \textbf{Step 2: Identify Activity Clusters}
        \STATE actClusters $\gets$ findActClusters(smoothActs) 
        
        \STATE \textbf{Step 3: Refine GPS Data}
        \STATE refinedGPS $\gets$ refineGPS(actClusters, trajectory) 
        
        \STATE \textbf{Step 4: Segment GPS Data}
        \STATE finalSegs $\gets$ segmentGPS(refinedGPS, actClusters) 
        \RETURN finalSegs
        
    \end{algorithmic}
\end{algorithm}

\textbf{Navigation}:
Leveraging crowdsourced trajectories and accurate segmentation results, we offer more precise recommendations for food pick-up and drop-off points. For example, in buildings where couriers lack access and must use dedicated lockers for food storage, the delivery destination provided by the customer and the designated drop-off location are not aligned well. In such scenarios, providing optimized drop-off point recommendations improves courier efficiency and reduces delays, particularly for new couriers.
\subsection{Elevation Change Detection}\label{sec:app:elevation}
We also explore another valuable application of activity recognition using the pretrained LIMU-BERT model: detecting elevation changes in couriers' movements. This enables the identification of more fine-grained activity statuses, such as whether couriers move vertically using an elevator or stairs. A straightforward solution is to use a barometer to measure air pressure and estimate elevation changes over time based on the barometric formula \cite{andrews2010introduction}. However, our analysis reveals that only 8\% of Android phones are equipped with barometer sensors, making this approach impractical for most devices. To overcome this limitation, we propose a general elevation change classifier with IMU sensors only, which can accommodate a broader range of smartphones.

\begin{figure}[t!]
  \centering
  \includegraphics[width=0.95\linewidth]{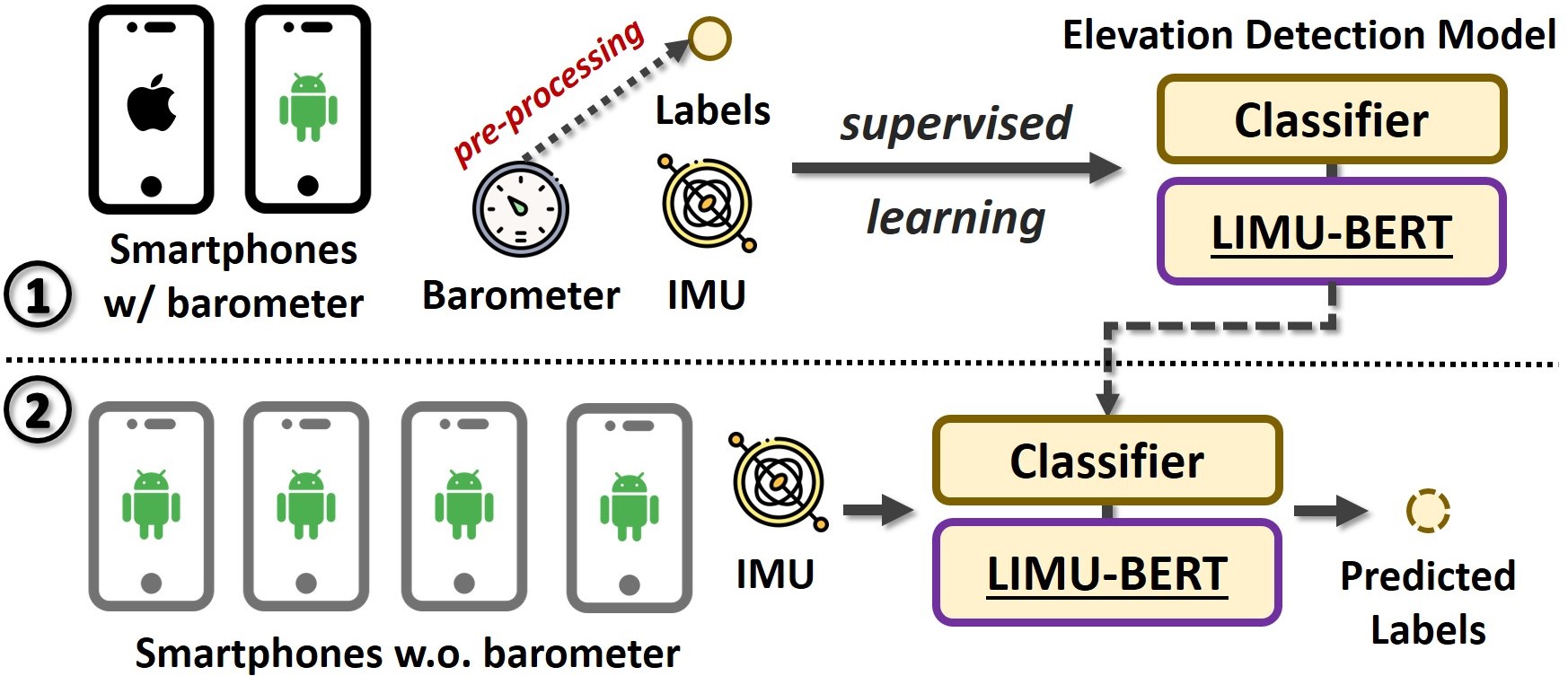}
  \caption{Elevation change detection design.}
  \label{fig:elevation}
\end{figure}

Although most Android phones lack barometer sensors, the majority of iOS devices are equipped with them. To exploit this opportunity, we propose a device collaboration training approach, as illustrated in Figure \ref{fig:elevation}. In the first learning phase, we utilize barometer data from smartphones equipped with barometers to generate two-class labels: vertical and non-vertical movements. Vertical movement is defined as an air pressure change of at least 0.25 hPa, with an overall change speed (air pressure change in hPa divided by time in seconds) exceeding 0.016 hPa/sec. These labels, along with the corresponding IMU data, are then used to train an elevation detection model based on the pretrained LIMU-BERT. In the second learning phase, the trained model is deployed on smartphones without barometers to detect vertical movement using IMU data alone. This collaborative approach bridges the gap between devices with and without barometer sensors.

We adopt the LIMU-BERT model pretrained in Phase II and fine-tune it using approximately 700K labeled samples. The model achieves an accuracy of 82.5\% and an F1 score of 82.4\% when evaluated on a dataset containing about 97K samples. It is important to note that when a courier is using an elevator moving at a constant speed, the vertical movement becomes indistinguishable from IMU sensor data. This limitation slightly lowers the overall performance of the vertical movement detection model compared to the 3-class activity recognition model. However, the overall performance is good and this application underscores the versatility of LIMU-BERT in adapting to diverse tasks. The vertical movement detection results, combined with activity status, provide comprehensive insights into couriers' behaviors.


\subsection{Estimated Time of Stop (ETS)}
Using accurate trajectory segmentation results, we can precisely identify when couriers get on and off their scooters and compute the "time of stop" — the duration from when a courier gets off the scooter to when they get back on. The "time of stop" is a critical metric for estimating overall delivery times and optimizing order assignment strategies. By analyzing large-scale "time of stop" data derived from trajectory segmentation, we develop a more robust model for  ETS.

\begin{table*}[h!]
\centering
\caption{Improved predictions in Estimated Time of Stop (ETS). A difficult area of interest (AOI) refers to an area where deliveries are challenging due to factors such as long walking distances or prolonged elevator wait times.}
\begin{tabular}{lccc}
\toprule
\multicolumn{1}{c}{\textbf{Category}} & \textbf{MAE Reduction (s)} & \begin{tabular}[c]{@{}c@{}}\textbf{Under-estimation}\\ \textbf{ Rate Reduction (\%)}\end{tabular}  & \begin{tabular}[c]{@{}c@{}}\textbf{Over-estimation}\\ \textbf{ Rate Reduction (\%)}\end{tabular} \\ \midrule
Overall & 1.8 & 4.98 & 4.57 \\ \midrule
Walk-only AOI & 3.6 & 3.92 & 9.34 \\ \midrule
Difficult AOI & 2.4 & 16.08 & 5.43 \\ 
\bottomrule
\end{tabular}
\end{table*}

In conducting our experiment to evaluate the impact of the adoption of LIMU-BERT on ETS, we construct a comprehensive dataset from major urban areas in China, including cities such as Shenzhen, Xi'an, and Guangzhou. The dataset consists of 2.38 million orders in the training set, 1.03 million in the validation set, and 780 thousand in the test set, all randomly selected from the period spanning mid-May to early June 2024. The focus is on measuring the accuracy of the estimated time of stop (ETS), our key performance indicator, by analyzing the Mean Absolute Error (MAE) reductions across various urban delivery scenarios. We also particularly examine long-tail orders, which are deliveries that fall outside typical patterns due to unpredictable factors or challenging logistics, as improvements in these predictions directly enhance customer satisfaction. The results are significant: overall MAE reduces by 1.8 seconds, with reductions in under-estimation and over-estimation rates by 4.98\% and 4.57\%, respectively. In pedestrian-only areas, we see a 3.6-second decrease in MAE, with more substantial declines in inaccurate estimations, while difficult delivery areas benefit from a 2.4-second MAE reduction. These findings demonstrate LIMU-BERT's ability to refine ETS predictions and thus ETA (visible to customers and couriers), which ultimately boosts customer satisfaction.

\subsection{Difficulty Analysis and Pricing Strategy}
Accurate trajectory segmentation results combined with vertical movement patterns, allow us to assess the challenges of specific delivery tasks. For example, if a restaurant is located on the upper floor of a complex shopping mall, couriers may face extended walking and waiting times for elevators. We will mark the restaurant location as "difficult". Similarly, if an area of interest (AOI) contains many such "difficult" restaurants or customer locations, the entire AOI is labeled as "difficult". This difficulty analysis is used to optimize pricing strategies, ensuring fairness by relatively increasing delivery fees for challenging AOIs and decreasing them for easier deliveries. This dynamic pricing approach establishes a fairer system that benefits both couriers and customers.

We conduct an A/B test experiment to evaluate a new pricing strategy designed to enhance the relative pricing of orders that require more effort to deliver, thereby improving the fairness of delivery compensation. This experiment is conducted in Shanghai from July 8, 2024, to July 14, 2024, with the experimental group comprising 510,000 orders and the control group comprising 2.02 million orders. We focus on two key variables: average delivery fees and the order acceptance rate. The findings indicate that the new pricing strategy reduces the average basic delivery fee per order by 0.06 yuan while maintaining the order acceptance rate within the first five minutes. From a business perspective, this reduction translates into considerable savings for the platform. \textit{With approximately 20 million orders processed daily, the average fee decrease of 0.06 yuan per order results in an estimated annual savings of around 0.44 billion RMB.} This significant cost-saving underscores the financial viability and efficiency gains from implementing a pricing strategy that balances operational demands with fair compensation, ultimately benefiting both the delivery platform and its stakeholders.



\section{Lessons Learned}
In our large-scale deployment of HAR models enabled by LIMU-BERT, several key lessons were gleaned from translating research into commercial applications.

\subsection{Scaling Law}
\begin{figure}[t!]
\begin{subfigure}{.47\linewidth}
  \centering
  \includegraphics[width=1.0\linewidth]{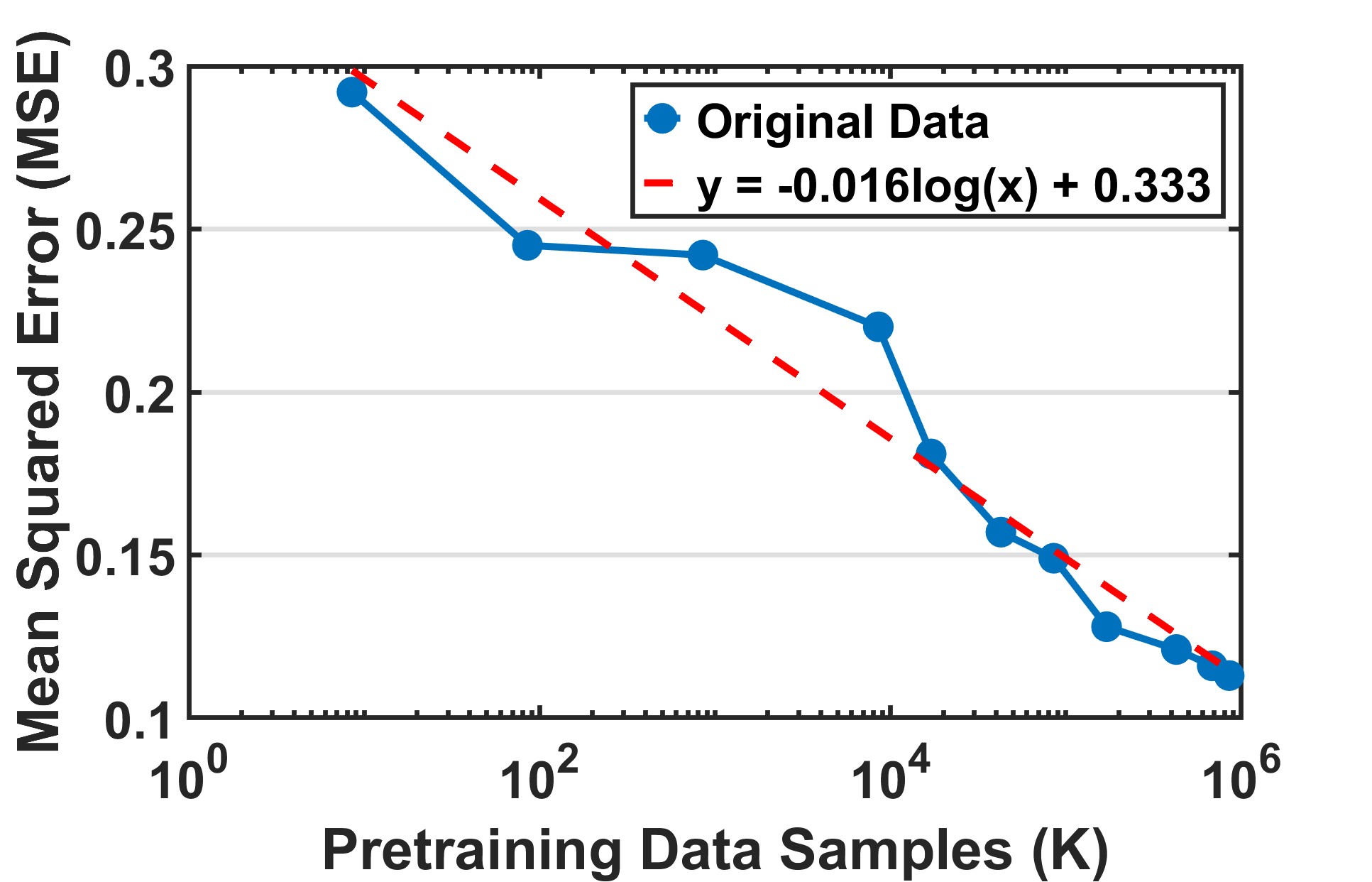}
  \caption{Data size}
  \label{fig:scale:data}
\end{subfigure}
\begin{subfigure}{.47\linewidth}
  \centering
  \includegraphics[width=1.0\linewidth]{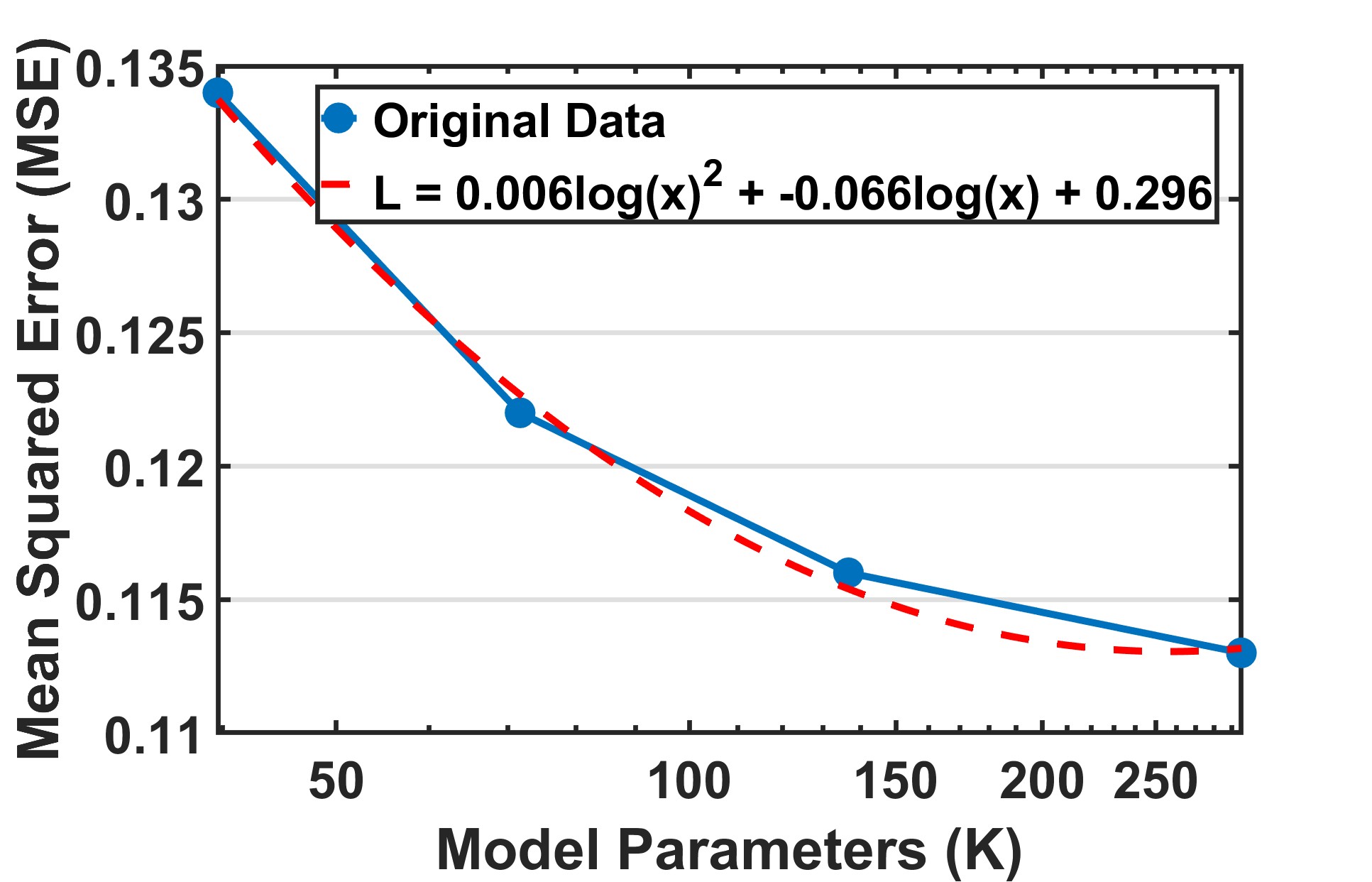}
  \caption{Model size.}
  \label{fig:sclae:model}
\end{subfigure}
\caption{Scaling performance of LIMU-BERT. The red line indicates the fitted curve for losses.}
\label{fig:scale}
\end{figure}
From our experiments, a key insight is the critical role of leveraging large-scale unlabeled data to enhance model performance. During the pretraining phase of LIMU-BERT, we sample an extensive unlabeled dataset collected nationwide from couriers, with a smooth and scalable data collection process. This highlights the strategic advantage of large-scale unlabeled data in real-world applications.

To further evaluate its impact, we conduct a scaling analysis of LIMU-BERT using up to 1.43 million hours of IMU data from over 50,000 couriers. As shown in Figure \ref{fig:scale}, the MSE loss steadily decreases as the number of training samples and model parameters increases. However, we observe a saturation effect, where increasing the model size does not guarantee that the loss approaches zero, and large model size also introduces higher computational overhead for mobile devices, presenting a trade-off. Hence, we utilize a dataset comprising 852 million samples and a model with 137 thousand parameters. Our findings confirm that the scaling law \cite{kaplan2020scaling} also applies to IMU foundation models, demonstrating an effective approach to improving the generalization of sensing models.

\subsection{Evaluating Models at Large Scale}
Another key lesson is the importance of having ground-truth labels to evaluate a model’s effectiveness in large-scale scenarios. Vision-assisted approaches can provide highly accurate labels; however, they are time-consuming and expensive. We employed external contractors to annotate approximately 420 hours of video footage for the Yangzhou dataset, which required over two weeks and cost more than 100,000 RMB. Such an approach is not scalable for nationwide deployment. To overcome this, we adopt a rule-based approach to generate labels using extensive data from other sensors and sensing techniques. We utilize IODetector \cite{zhou2022experience} and GPS speed to derive labels from large-scale unlabeled data. While this method may introduce noisy labels due to errors in IODetector or GPS speed measurements, it is cost-effective and scalable. Our results reveal that those crowd-sourced labels are useful and crucial for evaluating model effectiveness in real-world and large-scale scenarios.

\subsection{Handling Diverse Devices}
Commercial applications demand meticulous design and careful consideration to accommodate the diversity of smartphones with varying sensors and performance capabilities. One critical insight is that sensor availability can significantly impact system coverage and performance. During the initial phase of nationwide deployment, our monitoring system revealed that overall coverage of LIMU-BERT on couriers was at 89\%. Further analysis shows that approximately 57,000 devices were excluded from coverage due to the absence of gyroscope data or the lack of a gyroscope sensor. To address this gap, we develop an alternative model, \textit{accelerometer-only LIMU-BERT}, tailored for these devices. Specifically, we pre-train LIMU-BERT from scratch using accelerometer data and fine-tune it with the labeled samples from the Yangzhou dataset. This accelerometer-only model achieves a performance accuracy of 87.4\% and an F1 score of 86.3\% for the three-class activity recognition. The integration of this additional model increased courier coverage to approximately 99\%. Similarly, as detailed in Section \ref{sec:app:elevation}, the lack of a barometer in most smartphone models prompted us to design a universal, IMU-based model for detecting elevation changes. Both approaches ensure compatibility across a broader range of devices, addressing the limitations posed by hardware diversity.


Additionally, the choice of a 10 Hz sampling frequency considers not only model performance and transmission overhead but also device-specific factors. Even at this relatively low sampling rate, we observe that sensor data from certain devices are not always updated simultaneously, leading to synchronization issues and data gaps that can negatively affect model performance. To address this, we design a synchronization component to ensure all sensor data are properly aligned, allowing for consistent training and evaluation of models across a wide range of device conditions.

We also analyze the success rates of our SDK and find that most models achieve success rates above 99\%, and the few models with success rates below 90\% are primarily from less common brands, such as TECNO and Cancro. A likely reason is their lower system compatibility compared to standard Android distributions, which may impact the stability and functionality of the SDK.


\subsection{Sensor and Device Collaboration}
Although our primary goal is to build two IMU-based models, developing the entire system requires collaboration with other sensors. For crowdsourced labeling, we rely on GPS speed and IODetector \cite{zhou2022experience}, which incorporates data from light and magnetometer sensors. For the elevation change detection model, we utilize barometer readings to identify vertical movements, generating labels to train the IMU-based models. Additionally, device collaboration is sometimes necessary due to hardware limitations. Since most Android smartphones lack barometer sensors, we leverage labels derived from iOS devices that are typically equipped with barometers, to train models for most Android smartphones. This collaborative approach ensures broader applicability and robustness across diverse devices.

Another lesson learned is the recognition that certain behaviors or movements cannot be fully distinguished using IMU data alone. For example, as we observe in the confusion matrix displayed in Figure \ref{fig:cm:1}, a significant portion of "still" activities are erroneously classified as "riding" (8.6\%), and "walking" activities are often misclassified as "still" (8.0\%). This is because in certain scenarios, such as when a courier rides on flat terrain with constant speed and minimal maneuvers, the IMU pattern closely resembles that of standing still. Similarly, subtle movements during walking might resemble a "still" state when the courier is actively typing. This limitation necessitates exploring hybrid approaches that integrate additional sensor data (e.g., GPS) to discern between complex behaviors.

\subsection{Others}
During the deployment of the trained model, we discovered that directly compiling it using existing open-source tools, e.g., Pytorch Lite \cite{paszke2019pytorch}, resulted in a large model size, posing a critical issue for commercial applications due to app size constraints. We address this with a dynamic loading strategy, where only necessary packages are loaded at runtime as needed. It significantly reduces the app size, making the solution more efficient and practical for commercial adoption.

We also explore data augmentation techniques \cite{saeed2019multi, tang2021selfhar, chang2020systematic, xu2023practically}, specifically random scaling and rotation transformations, to enrich data diversity and thus enhance model performance. However, the results indicated that these augmentations only introduce marginal improvements: the accuracy slightly decreases by 1.24\%, while the F1 score increases by 1.74\%. While data augmentation remains a valuable method for enhancing model performance in general, its effectiveness may diminish when ample training data is extensive in our case. Future efforts should carefully consider the trade-offs and benefits of different augmentation techniques in the context of available data and specific use cases. 
\section{Discussion}
Since all orders are managed and fulfilled via smartphones in the on-demand food delivery service, couriers need to interact with their devices frequently and over extended periods. This continuous interaction introduces several concerns:

\textbf{Privacy concerns.}
To enable efficient order assignment and coordination, the platform relies on privacy-sensitive data such as location. But this data sharing ultimately benefits couriers by improving their working efficiency and income, while also enhancing the service quality of the platform. Couriers agree to share this data during their working hours as part of their contractual agreement. 

\textbf{Potential generalization.}
The working patterns of couriers are distinct from those in other professions, which affects the nature of the sensor data collected. As a result, the generalization of our model to other domains remains uncertain and may require fine-tuning. Nevertheless, based on our experience, LIMU-BERT offers a strong foundation, and adapting it to new domains typically requires only a modest amount of labeled data and training time.

\textbf{Future work.}
We aim to explore more applications of activity recognition on our platform to further enhance efficiency. For example, detecting the use of elevators or lifts in buildings can support more accurate time estimation and difficulty analysis in the delivery process.

\section{Related work}
\textbf{Real-world mobile system experience}. Recent studies \cite{kim2017smarter,hu2022experience,li2021experience,ni2022experience,alay2017experience,zhou2022experience,boateng2019experience,sevilla2019experiences,ding2021nationwide} have reported experiences with large-scale mobile system deployments. For example, Boateng et al. \cite{boateng2019experience} share their experience implementing wrist-worn computing devices for mobile health applications. Several other works \cite{ding2021nationwide,zhou2022experience,ni2022experience,hu2022experience} discuss large-scale location-related applications on mobile devices, such as indoor localization \cite{ni2022experience,hu2022experience} and indoor-outdoor detection \cite{zhou2022experience}. However, unlike existing systems that predominantly rely on signal processing methods, we are the first to present the nationwide deployment of deep learning-based models on mobile devices for human activity recognition. This work represents a milestone in the evolution of mobile systems in the era of artificial intelligence, showcasing the practical adoption of deep learning at scale.

\textbf{Activity recognition}.
Numerous efforts have been dedicated to activity recognition, resulting in a wide range of solutions. Compared to image-based or wireless approaches, wearable-based solutions are more ubiquitous and cost-effective. Previous studies \cite{yang2015deep,jiang2015human,yao2017deepsense,liu2020giobalfusion} have utilized supervised learning methods to classify activities using IMU sensor data. However, these methods heavily rely on large volumes of labeled data for training. Inspired by the success of self-supervised learning techniques, several studies \cite{saeed2019multi,xu2021limu, logacjov2024self, tang2021selfhar,qian2022makes,wang2022sensor,kara2024phymask,hong2024crosshar,narayanswamy2024scaling,hoddes2025scaling,sheng2022facilitating,dai2024contrastsense} have proposed building foundation models that leverage inexpensive and easily accessible data. These models can be adapted to various tasks, including activity recognition. In particular, we adopt LIMU-BERT \cite{xu2021limu}, a state-of-the-art foundation model for IMU data. Unlike the original work, which was trained on limited datasets, we construct LIMU-BERT using large-scale datasets and present its nationwide deployment for the food delivery service industry.

\textbf{On-demand food delivery.} 
On-demand food delivery has become a booming industry, primarily driven by advancements in technology and logistics. Alongside industry growth, numerous research efforts have explored how mobile sensing technologies can enhance the efficiency of food delivery services. Most works \cite{zhou2022experience,liu2022pred,guo2022wepos,ding2021nationwide,ding2022p2} focus on extracting location contexts of couriers, such as detecting indoor status or achieving merchant-level localization. Other studies \cite{dai2023opti,zhu2020order} address time inference for order servicing, including estimating the preparation time of orders. To the best of our knowledge, this is the first work to deploy nationwide activity recognition for on-demand food delivery, representing a significant advancement in leveraging human activity recognition to improve operational efficiency at scale.

\section{Conclusion}
This paper presents our experience in adopting human activity recognition (HAR) technology to support the real-world business of on-demand food delivery. We share our deployment experience, the lessons learned, and the practical considerations for adopting research innovations to large-scale commercial applications, paving the way for future advancements in mobile computing technologies.
\begin{acks}
We thank all reviewers for their insightful comments. This work is supported by the Global STEM Professorship Scheme of Hong Kong and the HKUST start up grant. Mo Li is the corresponding author.
\end{acks}

\balance
\bibliographystyle{ACM-Reference-Format}
\bibliography{sample-base}








\end{document}